\documentclass[pdflatex,sn-mathphys]{sn-jnl}% Math and Physical Sciences Reference Style
\usepackage{mathtools}

\algnewcommand\algorithmicforeach{\textbf{for each}}
\algdef{S}[FOR]{ForEach}[1]{\algorithmicforeach\ #1\ \algorithmicdo}
\jyear{2022}%

\theoremstyle{thmstyleone}%
\theoremstyle{thmstyletwo}%

\theoremstyle{thmstylethree}%

\begin{document}

\title{Persian topic detection based on Human Word association and graph embedding}

%TDHWA: a Topic Detection method on Persian social media stream using Human Word Association}
%1.	TDHWA: a topic detection method on Persian microblog (stream) using Human Word Association and Graph Embedding
%2.	Human Word Association coupled with Graph Embedding method on Persian microblog (stream) for topic detection
%3.	topic detection method on Persian microblog (stream) through(with) Human Word Association and Graph Embedding
%4.	Persian microblog’s Topic detection through(with) (using) Human Word Association and Graph Embedding
%5. Human Word association based (method for) topic detection on Persian microblog using graph embedding 
%6. Human Word association based graph embedding for topic detection on Persian microblog
%7. (ParsTD:) Persian topic detection (framework) based on Human Word association and graph embedding 


\author[1]{\fnm{Mehrdad} \sur{Ranjbar-Khadivi}}\email{mehrdad.khadivi@iaushab.ac.ir}

\author[1]{\fnm{Shahin} \sur{Akbarpour}}\email{akbarpour@iaushab.ac.ir}

\author*[2]{\fnm{Mohammad-Reza} \sur{Feizi-Derakhshi}}\email{mfeizi@tabrizu.ac.ir}

\author[1]{\fnm{Babak} \sur{Anari}}\email{anari@iaushab.ac.ir}

\affil[1]{\orgdiv{Department of Computer Engineering}, \orgname{Shabestar Branch, Islamic Azad University}, \orgaddress{\city{Shabestar}, \state{East Azerbaijan}, \country{Iran}}}

\affil*[2]{\orgdiv{Computerized Intelligence Systems Laboratory, Department of Computer Engineering}, \orgname{University of Tabriz}, \orgaddress{\street{29 Bahman Blvd.}, \city{Tabriz}, \state{East Azerbaijan}, \country{Iran}}}

\abstract{
This paper proposes a framework for topic detection in social media based on Human Word Association (HWA). Identifying topics within these platforms has become a significant challenge. Although substantial work has been done in English, much less has been conducted in Persian, especially for Persian microblogs. Furthermore, existing research primarily focuses on frequent patterns and semantic relationships, often neglecting the structural aspects of language. Our proposed framework utilizes HWA, a method that imitates mental capabilities for word association. This approach also calculates the Associative Gravity Force (AGF), demonstrating how words relate. Using this concept, a graph can be generated from which topics can be extracted through graph embedding and clustering. We applied this approach to a Persian language dataset collected from Telegram and conducted several experimental studies to evaluate its performance. The Experimental results indicate that our method outperforms other topic detection approaches.
}


\maketitle

%%%%%%%%%%%%%%%%%%%%%%%%%%%%%%%%--Section--%%%%%%%%%%%%%%%%%%%%%%%%%%%%%%%%%%%%%%%%%%%%
\section{Introduction}\label{sec:introduction}
\begin{figure}
    \centering
    \includegraphics[width=0.8\textwidth]{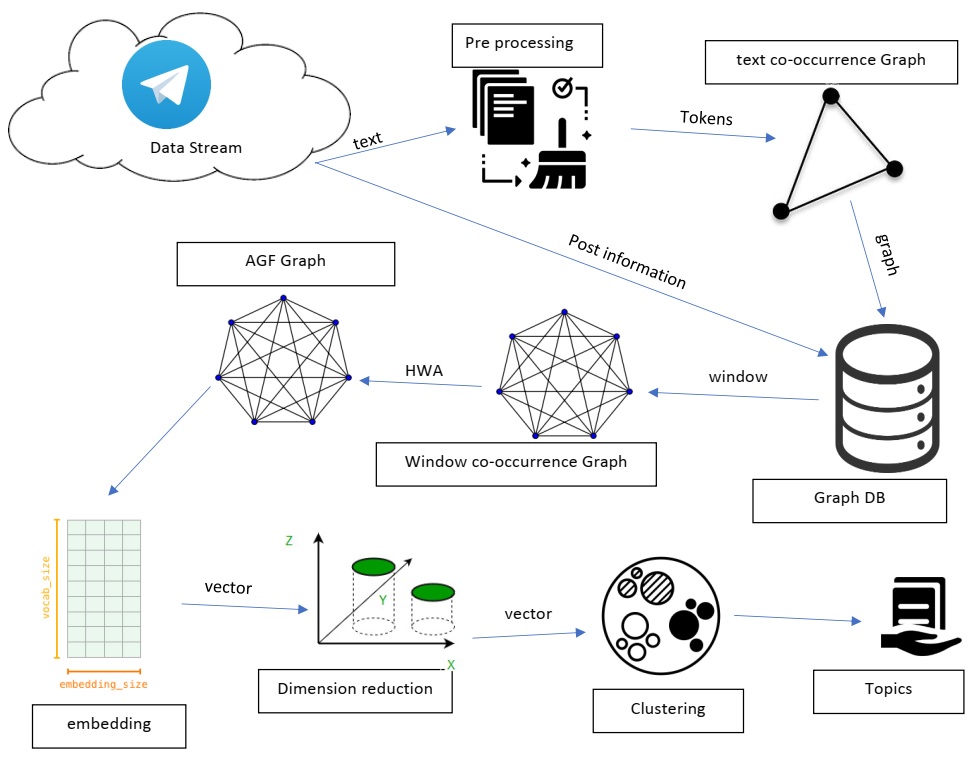}
    \caption{General structure of the proposed framework. this framework consists of different phases and steps, such as pre-processing, graph generation, embedding, and clustering.}
    \label{fig:General structure}
\end{figure}

Analyzing social media posts can help understand public events and people's opinions, concerns, and expectations \cite{COVIDTrendingTopics}. Identifying and examining social media topics aids in understanding the nature of emerging fields. This paper proposes a framework for topic detection based on Human Word Association (HWA). We used various combinations of graph embedding methods and clustering algorithms to determine the optimal approach. Initially, a stream of social media posts was captured. These posts' word co-occurrence was then obtained as a second step for HWA detection. This calculation led to a co-occurrence graph, to which the HWA-based AGF (Associative Gravity Force) formulation was applied, resulting in an AGF graph. Topics were then extracted using clustering algorithms. Each cluster could contain different topics. To achieve this, it was necessary to convert the graph into vector space, for which graph embedding methods were employed. Given the high dimensionality of this vector space, a dimension reduction method named UMAP was incorporated into the framework. Figure \ref{fig:General structure} presents the general structure of our proposed framework. When compared to other topic modeling methods, our framework demonstrated superior performance.
Here are the main contributions of this work:
\begin{itemize}
  \item Imitation of human mental ability for word association to extract topics in social networks
  \item Use of keyword co-occurrence for topic detection in social media messages
  \item Use of graphs to model the framework
  \item Development of a framework to extract topics from posts in the Persian language
\end{itemize}

%%%%%%%%%%%%%%%%%%%%%%%%%%%%%%--Sub-Section--%%%%%%%%%%%%%%%%%%%%%%%%%%%%%%%%%%%%%%%%%%
\subsection{Human Word Association}\label{subsec:HWA}
An intuitive method for analyzing Human Word Association (HWA) is to conduct surveys. These surveys are standard in psychology and linguistics, with the most popular being the free association test (FAT) and the free association norm (FAN). In these tests, participants are asked to write the first word that comes to mind in response to a given stimulus word. Studies have shown that Human Word Association is asymmetric, meaning the association strength between two words can vary depending on their order. Nelson et al. conducted a test with 6,000 participants and found, for example, that while the tuple (good, bad) is symmetric (since about 75\% of the participants answered 'good' for the word 'bad' and about 76\% wrote 'bad' in response to 'good'), the tuple (canary, bird) is asymmetric. Approximately 69\% of people answered 'canary' for the word 'bird', but only 6\% responded with 'bird' when presented with 'canary' \cite{Nelson2000} \cite{Nelson2004} \cite{Klahold2014}. Therefore, the proposed method must account for this asymmetry to simulate HWA accurately. In this paper, we employ a formulation named AGF for this purpose.

%%%%%%%%%%%%%%%%%%%%%%%%%%%%%%--Sub-Section--%%%%%%%%%%%%%%%%%%%%%%%%%%%%%%%%%%%%%%%%%%
\subsection{Graph Embedding}\label{subsec:graph_embedding}
Graph embedding methods fall into three standard categories: factorization-based, random walk-based, and deep learning-based \cite{Goyal2018}. This paper utilizes random walk-based methods, specifically the Deep Walk and Node2Vec models. These methods are advantageous when the graph is sufficiently large and cannot be fully measured \cite{Goyal2018}. We evaluated the results to choose the superior approach.

\begin{itemize}
\item \textbf{Deep walk} \cite{DeepWalk} involves the probability of observing the last k node and the subsequent k node of a random walk (RW) centered on the i-th vertex, as illustrated in equation  \ref{eq:1}.
    
\begin{equation}
    \label{eq:1}
    if \ len(RW)\ =\ 2k + 1\ \rightarrow\ max(logPr(\nu_{i-k},...,\nu_{i-1},\nu_{i+1},...,\nu_{i+k} \mid {Y_i})
\end{equation}
The model generates several random walks, each of length 2k + 1, to perform optimizations over the sum of the log-likelihoods of each random walk. Edges are then reconstructed from node embedding using 
dot-product-based decoders \cite{Goyal2018}.
\item \textbf{Node2Vec} \cite{node2vec} maintains high-order accessibility between nodes by maximizing the likelihood of subsequent nodes appearing in a random walk of fixed length. The main distinction is that Node2Vec uses a distorted random walk, balancing between breadth-first search (BFS) and depth-first search (DFS) graph searches. As a result, Node2Vec provides higher quality and more informative embeddings than Deep Walk. By striking the right balance, Node2Vec can preserve community structure and structural equivalence between nodes \cite{Goyal2018}.
\end{itemize}
%%%%%%%%%%%%%%%%%%%%%%%%%%%%%%--Sub-Section--%%%%%%%%%%%%%%%%%%%%%%%%%%%%%%%%%%%%%%%%%%
\subsection{Dimension Reduction}\label{subsec:dimension_reduction}
\begin{figure}
    \centering
    \includegraphics[width=0.8\textwidth]{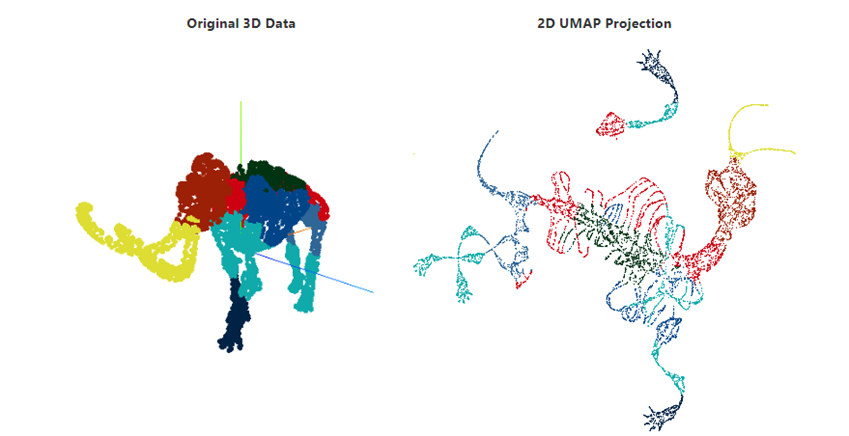}
    \caption{UMAP projections of a 3D woolly mammoth skeleton \cite{understanding-umap}}
    \label{fig:UMAP}
\end{figure}

Dimensionality reduction is a powerful tool for machine learning practitioners dealing with large and high-dimensional datasets. 
\begin{itemize}
\item \textbf{t-SNE} \cite{tsne} is one of the most widely utilized techniques for visualization. It is well-suited for visualizing high-dimensional datasets and employs graph-based algorithms for low-dimensional data representation. However, t-SNE does not scale well with rapidly increasing sample sizes, and attempts to speed it up lead to substantial memory consumption. Therefore, using t-SNE for clustering is not optimal. 
    
\item \textbf{UMAP} \cite{mcinnes2020umap} offers several advantages over t-SNE, including increased speed and better preservation of the data’s global structure. UMAP operates similarly to t-SNE by constructing a high-dimensional graph representation of the data, then optimizing a low-dimensional graph to match as closely as possible. Fig \ref{fig:UMAP} provides a visualization demonstrating the application of UMAP on a 2D projection of 3D data. Understanding the theory behind UMAP makes it easier to find out the parameters of the algorithm. The two most commonly used parameters, $n_{neighbors}$ and ${min}_{dist}$, effectively control the balance between local and global structures in the final projection.
\end{itemize}

%%%%%%%%%%%%%%%%%%%%%%%%%%%%%%--Sub-Section--%%%%%%%%%%%%%%%%%%%%%%%%%%%%%%%%%%%%%%%%%%
\subsection{Clustering for Topic Detection}\label{subsec:clustering_topic}
Clustering is a method used to identify topic communities. This paper evaluates the effectiveness of two different approaches, K-means, and HDBSCAN. 

\begin{itemize}
\item \textbf{K-means} is a well-known iterative clustering algorithm based on iterative relocation. It partitions a dataset into k clusters, minimizing the mean squared distance between the instances and the cluster centers \cite{TextClustering}.

\item \textbf{HDBSCAN} \cite{HDBSCAN} developed by Campello et al., is a clustering algorithm where DBSCAN requires a minimum cluster size and a distance threshold epsilon as user-defined input parameters. Unlike DBSCAN, HDBSCAN implements various epsilon values, requiring only the minimum cluster size as a single input parameter. The cluster selection method returns the cluster with the highest stability over epsilon. This approach allows for the identification of variable density clusters without the need to select an appropriate distance threshold.
\end{itemize}

The rest of this paper is structured as follows: Section \ref{sec:related_work} reviews related work in this field. Section \ref{sec:proposed_method} presents the details of the proposed framework. Section \ref{sec:exp-results} provides an overview of the dataset, evaluation metrics, and experimental results. Finally, Section \ref{sec:conclusion} concludes the paper.

%%%%%%%%%%%%%%%%%%%%%%%%%%%%%%%%--Section--%%%%%%%%%%%%%%%%%%%%%%%%%%%%%%%%%%%%%%%%%%%%
\section{Related Works} \label{sec:related_work}
This section examines various methods in the field of subject recognition in social networks. This operation is divided into two categories in some sources: document pivot and feature pivot \cite{Aiello2013,EnhancedHbGraph}. While other sources divide the feature pivot category into two categories: based on the feature and the probabilistic topic model \cite{Indra2018, HUPM, HUPC}. The document-based approach is a subject recognition technique that uses document clustering based on similarities. This technique was developed based on First Story Recognition (FSD) research. Feature pivot approaches cluster documents based on Feature selection. Determining the threshold value and probabilistic model are two methods for document feature selection. 
One of the threshold value determination approaches is the TF-IDF. In contrast, one of the features based on the probabilistic subject model is the document burst. The Burst document has a higher frequency of occurrence than other documents.
Most scientific achievements in the field of topic modeling are based on the Dirichlet latent allocation method (LDA) \cite{Aiello2013, Blei2003}. LDA is a probabilistic model built on BoW \footnote{Bag of Word} and widely used for subject modeling. Word repetition rates are extracted from documents to calculate the probability distribution of words that are likely to be found in topics.

Most topic modeling improvements are based on frequent pattern mining(FPM). This method has several applications in many fields, such as clustering and classification. Frequent pattern mining is a conventional method for detecting topics on Twitter data streams, and various frameworks are available \cite{Saeed2019}.

In \cite{SFPM}, a soft frequent pattern mining(SFPM) is designed to overcome the topic detection problem. This study aims to find all word co-occurrence with a value greater than two, so the probability of each word is calculated separately. After finding top-K frequent words, a co-occurrence vector is formed to add the words that occurred together to the top-K word vectors. As an improvement to this approach, in \cite{Gaglio2015}, by adding a named entity recognition weight boost to enhance word score, a system has been designed to overcome the limitation of SFPM in dealing with dynamic and real-time scenarios. A method called HUI-Miner has been introduced in \cite{Liu2012} to identify a set of high-utility itemsets without generating a candidate. This method has been used in \cite{HUPM, HUPC, Gaglio2015} to find high utility itemsets in the domain of topic detection on Twitter. \cite{HUPC} introduced a high utility pattern clustering method (HUPC). After determining each pattern's utility and extracting high utility patterns, this method selects top-K very similar patterns using the modularity-based clustering method and KNN classification method. Also, \cite{HUPM} uses a high utility pattern mining method (HUPM) to find a group of frequently used words, and then a data structure called TP-tree is used to extract the pattern of the main topic.
Saeed et al. \cite{EnhancedHbGraph, Saeed2019, Saeed2018, Saeed2020} introduced a dynamic heartbeat graph (DHG). This algorithm creates a subgraph for each sentence. These subgraphs are then added to the main graph and applied to the entire graph based on the degree of coherence of the words (edges between nodes) \cite{Asgari-Chenaghlu2021}. The methods that have been introduced so far are based on the frequency of patterns or the co-occurrence of words. 
The semantic relationship between words needs to be addressed in approaches. \cite{TopicBert} uses a combination of transformers with an incremental community detection algorithm to identify topics. On the one hand, transformers provide a semantic relationship between words in different contexts. On the other hand, the graph mining technique improves the accuracy of the resulting subjects with the help of simple structural rules \cite{TopicBert}. 
It should be noted that the topic detection problem is a natural language processing problem, so the linguistic approaches still need to be investigated. Therefore, language structural methods and text processing will improve the results. Khadivi et al. in \cite{HWA_Embedding_MRK} use Human Word association (HWA) as a structural language technique to imitate the human mental ability to associate words. They also employed embedding techniques for better results.
As previous researches show, It is better to illustrate social network mining using Graphs. This framework is a graph mining method that uses HWA.

%%%%%%%%%%%%%%%%%%%%%%%%%%%%%%%%--Section--%%%%%%%%%%%%%%%%%%%%%%%%%%%%%%%%%%%%%%%%%%%%
\section{Proposed Method} \label{sec:proposed_method}
\begin{figure}
    \centering
    \includegraphics[width=0.9\textwidth]{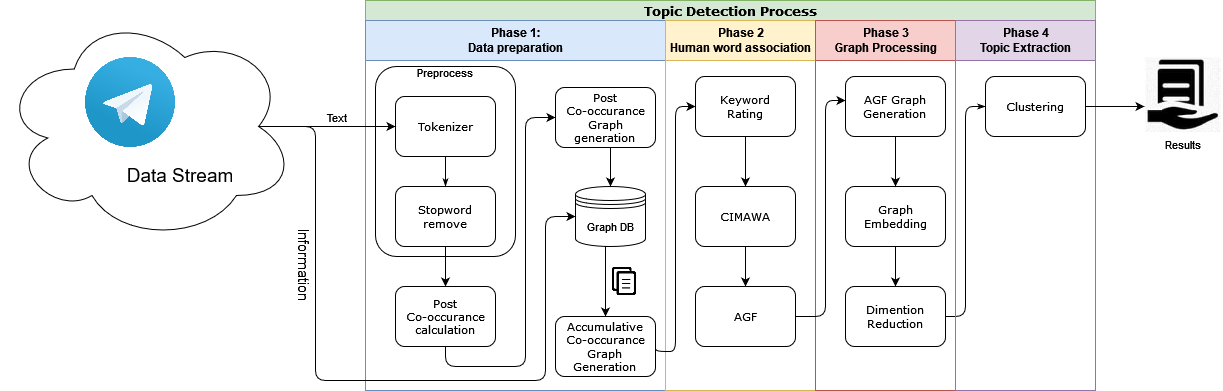}
    \caption{The flowchart of the proposed framework. This framework has four phases and thirteen steps in total. These steps are pre-processing, co-occurrence Graph generation, 3 steps for HWA, graph embedding, dimension reduction, and clustering. }
    \label{fig:flowchart}
\end{figure}

In principle, FPM methods cannot track the order of items as they are applied to transactions and itemsets. However, employing linguistic structure processing methods proves promising when dealing with social media posts, a form of natural language processing. One beneficial strategy to reinforce these methods involves understanding how humans interpret topics. A technical solution can be developed by concentrating on the human ability to identify associations between words. To comprehend association building, a concept known as HWA is utilized to extract topics that accurately represent the flow of data in social networks.
The social network data stream can be considered a sequence of posts received chronologically. All recently received posts within a fixed time window (for example, 12 hours) can be represented as a batch of posts like ${Batch}^L = \{ {Post}_p^L \}$ where ${Post}_p^L$ is, post $p$ in window $L$. It should be noted that in this paper, a post is the text of that post after pre-processing and removing stopwords. If $W^L = \{ w_l^L,w_2^L,...,w_\Omega^L \}$ be a set of words in the window $L$, it can be said that ${post}_p^L \subset W^L$. In other words, ${post}_p^L$ is a set of words; each word is a member of the word set of the same window. The final purpose of this paper is to extract topics from window $L$, which is presented as $T^L$. This article proposes a method with 4 phases to carry out this process; each phase has its steps. These phases and steps are described in the following. The flowchart of this phase and their steps are also shown in figure \ref{fig:flowchart}.

\subsubsection*{Phase 1: data preparation}

\begin{figure}
    \centering
    \includegraphics[width=0.3\textwidth]{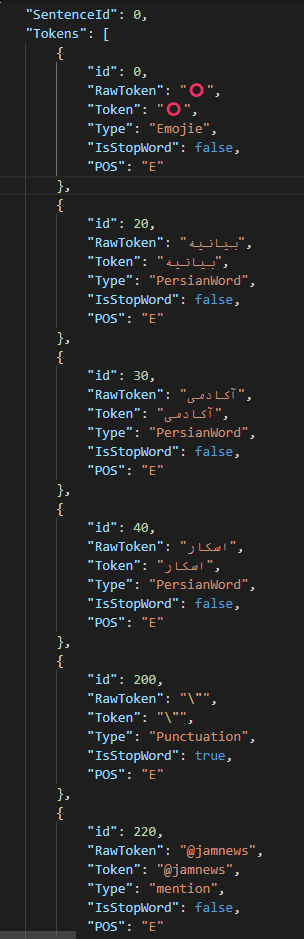}
    \caption{An example output of the tokenizer}
    \label{fig:tokenizer}
\end{figure}

\textbf{Step 1:} 
Posts on social networks are likely to include noise, such as typos, emojis, mentions, hashtags, or URLs. This noise could result in numerous issues; hence, it is essential to pre-process these posts to reduce the noise level. A tokenizer, specifically created for parsing social network posts, has been employed for this task. This tokenizer splits tokens based on different types of separators and identifies the token type. Examples of token types include emojis, URLs, hashtags, mentions, numerals, and simple or compound words. \ref{fig:tokenizer} provides an example of the output of this tokenizer. Subsequently, apart from numbers and words, all tokens of the post are eliminated, including stop words. A list of tokens processed by the tokenizer will proceed to the second step to calculate the co-occurrence.

\begin{figure}
    \centering
    \includegraphics[width=0.8\textwidth]{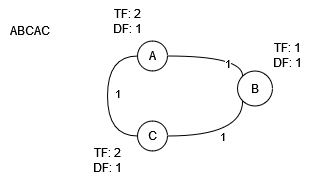}
    \caption{Post-Co-occurrence graph. If 'ABCAC' is the text of a post, The equivalent graph is shown in the figure.}
    \label{fig:G_CooC_Post}
\end{figure}

\textbf{Step 2:} A post, after pre-processing, is represented as ${Post}_p^L = \{ Word_{\omega}^L \}$. This post can also be illustrated using a graph. The co-occurrence graph of post $p$ is defined as ${Gcooc}_p^L(V, E)$. ‌The reason is, each vertex of the graph represents each word of the post, and the edge between two different vertices symbolizes a weight function, as shown in eq. \ref{eq:E_CooC}. To use this graph to generate another graph, a graph for batch $L$ (described later), Each node also has two properties $TF$ and $DF$. These properties are added to reduce the Computational complexity of generating that graph. As ${Post}_p^L$ is set, property $TF$ indicates the frequency of words.  Also, $DF$ represents the number of posts where the word occurred; for now, its initial value is 1. An example of this graph is represented in fig.\ref{fig:G_CooC_Post}.

\begin{equation}
        \label{eq:E_CooC}
        E_{{Gcooc}_p^L}(x,y) = {CooC}_p^L(x,y) 
    \end{equation}
    
where ${CooC}_p^L(x,y)$ is calculated as eq.\ref{eq:CooC}. If $x$ and $y$ are assumed to be two different words, and both have occurred in ${Post}_p^L$, then their co-occurrence is equal to 1; otherwise, it is 0;

\begin{equation}
        \label{eq:CooC}
        {CooC}_p^L(x,y) = \begin{cases} \mbox{1} & \mbox{if }  x \in {Post}_p^L  \&  y \in {Post}_p^L \\
                            \mbox{0} & \mbox{otherwise} \end{cases}                                    x \neq y \bigwedge x, y \in {W^L}
    \end{equation}

\begin{figure}
   \centering
    \includegraphics[width=0.8\textwidth]{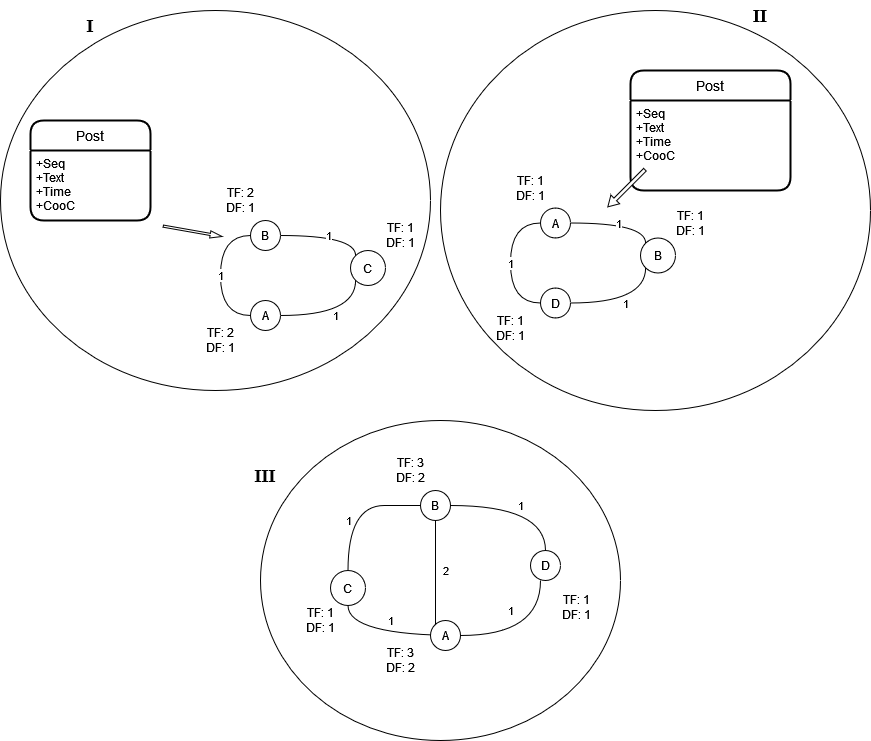}
    \caption{
    Window co-occurrence graph generation. If a window(ex. $B^L$) consists of 2 posts and $P_i$ and $P_j$ are the text of these posts, the text co-occurrence graph of each post is represented in I, II. III is a cumulative graph which is a Windows co-occurrence graph. I, II: The $TF$ tag of each node is assigned based on the frequency of the word, and the $DF$ tag also takes the initial value of 1. The values of the edges are also considered 1. III: This graph is formed by combining A and B. And the weight of the edges and nodes is updated using Algorithm \ref{alg:cooc_graph}}
    \label{fig:GCooC}
\end{figure}

As the data stream receives and new posts arrive, the newly captured post’s graph will be generated and added to the previous graph. A more extensive graph will be achieved at the end of the batch. This graph, named Batch CooC graph (BCG), is defined as ${Gcooc}^L(V^L, E^L, {TF}^L, {DF}^L)$. The difference between the Batch CooC graph and Post CooC graph is that in BCG, All nodes(V) are members of $W^L$, and edges, which are CooC(x,y), are calculated using eq. \ref{eq:CooC_tot}. An example of this process can be seen in Fig. \ref{fig:GCooC}.

\begin{equation}
        \label{eq:CooC_tot}
        E(x,y) = {CooC}^L = \sum_{p=1}^{\|{Batch}^L\|}  {CooC}_p^L(x,y)
    \end{equation}

Also, $TF$ and $DF$ tags are calculated using equations \ref{eq:TFW} and \ref{eq:DFW}, respectively.

\begin{equation}
    \label{eq:TFW}
    {TF}^L(w) = \sum_{p=1}^{\|{Batch}^L\|}  {TF}_p^L(w)
\end{equation}

\begin{equation}
    \label{eq:DFW}
    {DF}^L(w) = \sum_{p=1}^{\|{Batch}^L\|}  {DF}_p^L(w)
\end{equation}

The process of generating BCG and updating the weight of the edges and nodes properties is explained in Algorithm \ref{alg:cooc_graph}.

\subsubsection*{Phase2: HWA}\label{subsec:Phase2}

As the time window is completed and the graph generation process is over, the BCG graph becomes quite large in terms of vertices and edges, leading to computational complexity. Therefore, a solution is needed to prune the graph and remove extra information. One basic solution is to use linguistic-based methods, and the method employed in this article is based on HWA.

\textbf{Step 3:} As each vertex of the BCG graph represents a word, word ranking becomes the first step in identifying important and frequent words, known as keywords. This framework uses a method derived from the conventional TF-IDF method. ${Kr}^L(w)$ is formulated as eq. \ref{eq:kr}.

\begin{equation}
    \label{eq:kr}
    {Kr}^L(w) = {TF}^L(w) \times log\frac{\|{Batch}^L\|}{{DF}^L(w)}
\end{equation}

Words in $W^L$ with a high ${Kr}$ value are the most important words, referred to as Keywords. Deciding how many words to select requires a threshold, referred to as Rate in this paper. Choosing a fixed rate is incorrect, given that the number of words in each window varies. Therefore, h\% of words with high ${Kr}$ will be selected as keywords. Various experiments have been performed to achieve an optimal rate. Based on the parameter tuning in Table \ref{tbl:ParamSetup}, the best value is $h=10$. Therefore, words not in the top 10\% will be pruned from the $G_{cooc}$.

\begin{algorithm}
\caption{Generation of $G_{CooC}$}
\label{alg:cooc_graph}
\begin{algorithmic}[1]
\State \textbf{Input:} datastream
\State \textbf{Output:} $G_{CooC}$
\State $G_{CooC} \Leftarrow NULL$
\ForEach {post \textbf{in} ${Batch}^L$}
    \ForEach{$V_i$ \textbf{in} $G_i$}
        \If{$V_i$  \textbf{not in} $G_{CooC}$}
            \State Add $V_i$ to $G_{CooC}$ as $V$
            \State $V.TF = V_i.TF$
            \State $V.L_M = V_i.L_M$
        \Else
            \State $V.TF += V_i.TF$
            \State $V.L_M += V_i.L_M$
        \EndIf
    \EndFor
    \ForEach{$E_i$ \textbf{in} $G_i$}
        \State $x, y = nodes(E_i)$
        \If{There is no Edge between $x$ and $y$ in $G_{CooC}$}
            \State Add an Edge between $x$ and $y$ as $E$
            \State $E(x, y) = E_i.weight$
        \Else
            \State $E(x, y) += E_i.weight$
        \EndIf
    \EndFor
\EndFor
\end{algorithmic}
\end{algorithm}

\textbf{Step 4:} In this step, to replicate a human perception of HWA and the symmetric and asymmetric combination of word association, a concept called CIMAWA has been used, defined as eq. \ref{eq:CIMAWA}.

\begin{equation}
    \label{eq:CIMAWA}
    {CIMAWA}^L(x,y) = \frac{{CooC}^L(x,y)}{{TF}^L(y)} + \delta \times \frac{{CooC}^L(x,y)}{{TF}^L(x)}
\end{equation}

$\delta$ is a damping factor assumed to be 0.1 based on the experiment results. CIMAWA is used to calculate AGF in the next step.

\textbf{Step 5:} The Associative Gravity Force (AGF) quantifies the strength of the association of word x with word y in a text window. In other words, it calculates the gravitational pull of word $x$ concerning the word $y$. It should be noted that according to the CIMAWA concept, $AGF(x, y)$ will be different from $AGF(y, x)$. The function used in this paper to calculate the power of word association (AGF) is given in eq. \ref{eq:AGF}.

\begin{equation}
    \label{eq:AGF}
    {AGF}^L(x,y) = {CIMAWA}^L(x,y) \times \frac{{Kr}^L(x)}{{Kr}^L(y)}
\end{equation}
    
\begin{figure}
    \centering
    \includegraphics[width=0.8\textwidth]{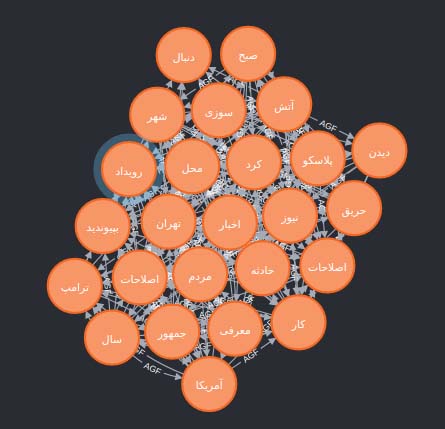}
    \caption{A sample of AGF graph. This graph represents each word as a node and edge weight is calculated using equation \ref{eq:AGF}}.
    \label{fig:AGF_graph}
\end{figure}

\textbf{Step 6:} By calculating the Associative Gravity Force, the previous parameters should be replaced with new parameters; in other words, a new graph will be created. Although this new graph has fewer parameters, these parameters provide more information about the type of relationship between vertices (words); Because they are generated using linguistic structures. This paper proposes a new graph structure. it is defined as $G_{AGF}$($V$, $E$) where $V$ are the keywords of $W^L$ as nodes, and $E$ is the weight of each directed edge called AGF, which can be calculated using phase 2. Unlike the CooC graph, the AGF graph is a directed graph. Fig \ref{fig:AGF_graph} shows a sample of the AGF graph.

\subsubsection*{Phase 3: Graph Processing}
Up until this point, posts and words have been converted into graphs. The goal is to identify their topics, and one of the best ways to interact with the social network graph is to find the community. Therefore, this graph must be processed in an understandable way to the algorithm. For this purpose, the graph embedding method has been used. The following phase will involve clustering.

\textbf{Step 7:} This paper suggests using Graph Embedding. Two methods, DeepWalk and Node2Vec, have been evaluated to identify a more accurate method. Different parameter values of these algorithms are tested to select the best algorithm, and the best values are set. The results of the parameter tuning, as well as the evaluation results of both algorithms in section \ref{subsec:ParamSetup}, show that Node2Vec performs best. These vectors will be clustered in the following phase to find communities, but the graph is very dense. This may reduce the accuracy of the topic extraction. One way to increase accuracy is to reduce the dimension.

\textbf{Step 8:} Dimensionality reduction is a powerful tool for machine learning practitioners to visualize and understand large, high-dimensional datasets. UMAP can be used as an effective preprocessing step to boost the performance of density-based clustering. As in the graph embedding step, the UMAP parameters are fully tuned. The main parameter is $N_{neighbors}$, and its value is considered 2 based on the experiments. The parameter tuning results are given in section \ref{subsec:ParamSetup}, and the evaluation results are given in section \ref{subsec:results}.

\subsubsection*{Phase 4: Topic Extraction}
Extracting topics from a social network graph is similar to finding communities in social networks. One way to detect communities is by applying clustering to the social network graph. As previously mentioned, a social network is a high-density graph. Also, topics in social networks have a hierarchical structure. Given these considerations, different clustering methods are evaluated in this paper.

\textbf{Step 9:} 
Once the embedding of each node is prepared, they are clustered. This paper evaluated two different clustering algorithms: K-means and HDBSCAN. The primary purpose of doing this is to find topics over a period of time. To find optimum parameters for each of the clustering algorithms, several experiments have been conducted. The selected values for these parameters are presented in Table \ref{tbl:ParamSetup} and discussed in \ref{subsec:ParamSetup}. The results of each method within the proposed framework are shown in section \ref{subsec:results}.

%%%%%%%%%%%%%%%%%%%%%%%%%%%%%%%%--Section--%%%%%%%%%%%%%%%%%%%%%%%%%%%%%%%%%%%%%%%%%%%%
\section{Experiments and Results}\label{sec:exp-results}
This section presents the experimental results obtained by the proposed framework. All implementations were done in Anaconda Python 3.8 and ran on a PC with a 3.60GHz Core i7-7700 processor with 32 GB RAM and a Linux Mint operating system. The performance of the proposed work is compared with each other and with the results of other models. All comparative models are also implemented and executed over the dataset introduced in \ref{subsec:dataset}, using evaluation metrics in section \ref{subsec:evaluation_metrics}.

%%%%%%%%%%%%%%%%%%%%%%%%%%%%%%--Sub-Section--%%%%%%%%%%%%%%%%%%%%%%%%%%%%%%%%%%%%%%%%%%
\subsection{Dataset}\label{subsec:dataset}
\begin{figure}
    \centering
    \includegraphics[width=0.9\textwidth]{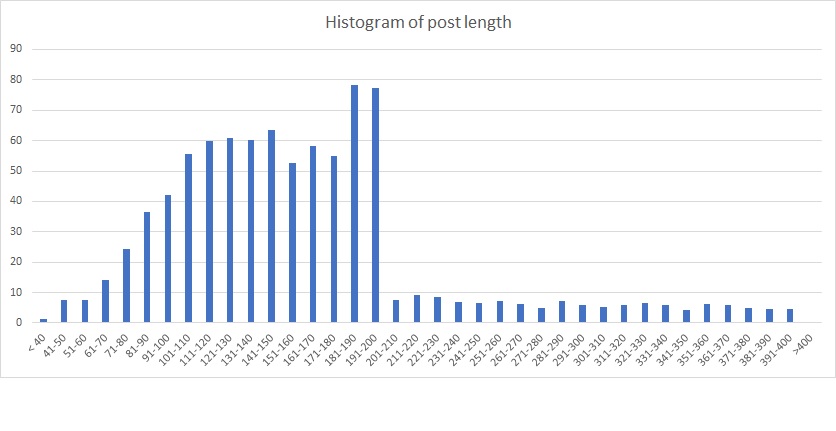}
    \caption{Histogram of Telegram posts' character length. This histogram shows that posts shared on Telegram have between 70 - 200 character lengths. This is the same as other microblog platforms like Twitter.}
    \label{fig:Histo}
\end{figure}

To evaluate the performance of the proposed algorithm, the Sep\_General\_Tel01 dataset \cite{Sep-TD-Tel}, provided by the Computerized Intelligence Systems Laboratory (ComInSyS)\footnote{\url{cominsys.ir}} is used. This dataset has been prepared due to the limited resources in Persian and the high popularity of the Telegram social network in Iran. The Telegram network is considered a microblog. A microblog is a small content designed to engage an audience quickly. Microblogs are short blog posts (less than 300 words) that can contain images, GIFs, links, infographics, videos, and audio. Figure \ref{fig:Histo} illustrates the histogram of the character length of Telegram posts.

\begin{figure}
    \centering
    \includegraphics[width=0.9\textwidth]{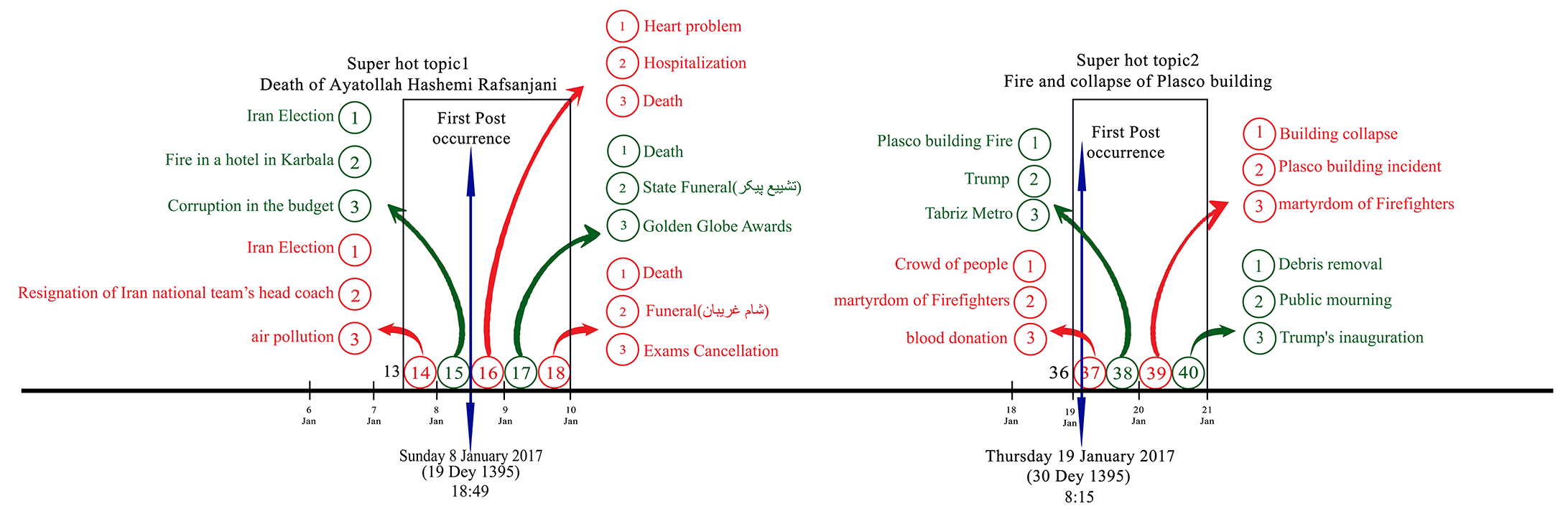}
    \caption{Dataset timeline. The horizontal axis shows the days from 1 January 2017 to 21 January 2017. Because the window size is 12 hours, each day is divided into 2 windows. Each of these batches contains different topics since only \{14, 15, 16, 17, 18\} and \{37, 38, 39, 40\} windows which are related to two super hot topics, "The death of Ayatollah Hashemi Rafsanjani" and "The fire in the Plasco building" respectively have grand-truth and are tagged by experts. For example, the Top-3 topics for window 14 are as follows: \{Iran election, the resignation of Iran national teams head coach, air pollution\}. It is important to note that the first post related to the "The death of Ayatollah Hashemi Rafsanjani" topic occurred on Sunday, January 8, 2017, at 18:49; in other words, the first post was published at this time. }
    \label{fig:DataSet}
\end{figure}

In this work, the official API published by Telegram is used. The implemented framework can fetch text messages and media (such as images, documents, and videos). In order to respect privacy principles, only data related to channels and public groups have been used for collection. Unlike existing datasets, which are often collected from the Twitter social network and require keywords to fetch information, the Telegram network does not need any. This dataset is also collected without using any keywords. This dataset contains more than ten thousand records of messages sent to public channels and groups in one month between 12 Dey 1395 (1 January 2017) and 12 Bahman 1395 (31 January 2017). This data includes two major topics: "The death of Ayatollah Hashemi Rafsanjani" and "The fire in the Plasco building". Due to the nature of this research, only the text of the messages sent has been used. To process posts, this database is divided into sixty 12-hour windows. To evaluate and compare the results of this paper, it is necessary to compare the extracted topics with GT. For this purpose, nine windows with the most news value and the best match with the two cloud themes have been labeled out of the sixty available windows. These windows are \{14, 15, 16, 17, 18, 37, 38, 39, 40\}. To perform this operation, four experts were used for labeling, and the final label was the result of the opinions of these individuals. A summary of information about this database to better understand the timeline of this database is given in Figure  \ref{fig:DataSet}.

%%%%%%%%%%%%%%%%%%%%%%%%%%%%%%--Sub-Section--%%%%%%%%%%%%%%%%%%%%%%%%%%%%%%%%%%%%%%%%%%
\subsection{Parameters Setup}\label{subsec:ParamSetup}
All of the parameters used in this framework are shown in Table \ref{tbl:ParamSetup}. This table represents the methods used in this paper, their parameters, their range, and the selected value for each one. Table \ref{tbl:ParamSetup} summarizes the parameters used in the framework. The framework has been executed with different values for each parameter and fixed ones for the others. Different methods have been used in this paper, and each method has different parameters. For example, $h$ and $\delta$ represent Keyword rating and damping factor, respectively, which are parameters used in HWA methods. Both parameters are tuned in various numbers to find a suitable value. For instance, parameter $h$ has an initial value of 5 and a termination value of 50. This parameter will increase by 5. Unlike this, some parameters have fixed values. For example, parameters like 'number of walks', which shows the number of walks to sample for each node, and 'walk length', which represents the length of each walk, have fixed values of 64 and 8, respectively. Besides that, some parameters have a list of values. Parameters $P$ and $Q$ in the Node2Vec method are of this type. Nevertheless, these parameters in DeepWalk are fixed.

\begin{table}[]
    \centering
    \caption{The parameter and their values used in the proposed method. Some of the parameters are tuned to get better results. For example, parameter $h$ from keyword rating is tuned over the range of [5, 50] with 5-step incrementation, and the selected value is 10. Some parameters are assumed to be fixed as they have little effects, like the number of walks, walk length, or max iteration. Also, parameters $P$ and $Q$ are used in both Node2Vec and DeepWalk methods, but they are not as Variable in DeepWalk as in Node2Vec.}
    
    \label{tbl:ParamSetup}
    \begin{tabular}{cccc}
        \hline
        Method & Parameters & Span & Value\\
        \hline
        \multirow{2}{*}{HWA} 
            & Keyword Rate(h)   & range( 5, 50, 5)     & 10 \\
            & Damping factor($\delta$ ) & range(0.1, 1.0, 0.1) & 0.1\\
        \hline
        \multirow{7}{*}{GE(Node2Vec)}
            & number of walks     & & 64 \\
            & walk length         & & 8  \\ 
            & dimensionality of the feature vectors(D)  & & 32 \\
            & window size(W)         & & 10 \\
            & number of iterations(epoch) & range(10,1000,10) & 100 \\
            & in-out parameter(p) & [0.25, 0.50, 1.0, 2.0, 4.0] & 0.50\\
            & return parameter(Q) & [0.25, 0.50, 1.0, 2.0, 4.0] & 1.0\\
        \hline
        \multirow{7}{*}{GE(Deepalk)} 
            & number of walks     & & 64 \\
            & walk length         & & 8  \\ 
            & dimensionality of the feature vectors(D)  & & 32 \\
            & window size(W)         & & 10 \\
            & number of iterations(epoch) & range(10,1000,10) & 100 \\
            & in-out parameter(p) & & 1\\
            & return parameter(Q) & & 1\\
        \hline
        \multirow{1}{*}{MAP} 
            & Number of neighbors &  & 2 \\
        \hline
        \multirow{2}{*}{Clustering (Kmeans)} 
            & Number of clusters & range(2,20,1)  & 8 \\
            & Max iteration &  & 300\\
        \hline
        \multirow{1}{*}{Clustering (Hdbscan)} 
            & Min cluster size & & 5\\
        \hline
    \end{tabular}
\end{table}

%%%%%%%%%%%%%%%%%%%%%%%%%%%%%%--Sub-Section--%%%%%%%%%%%%%%%%%%%%%%%%%%%%%%%%%%%%%%%%%%
\subsection{Evaluation Metrics}\label{subsec:evaluation_metrics}
As topic detection involve identifying groups or clusters, clustering evaluation is relevant in topic detection tasks. Cluster evaluation is the process of measuring the quality and performance of a clustering output. Cluster evaluation can help to find out how well the proposed framework works. If ground truth is available, it can be used by methods that compare the clustering against the ground truth and measure. "ground truth" is the labeling of the data, such as a human expert's judgment or a benchmark dataset. Two measures, F-measure and FS criterion, are adopted to evaluate the clustering quality. It is desirable to maximize the F-measure and minimize the FS criterion of clusters.

%%%%%%%%%%%%%%%%%%%%%%%%%%%%--Sub-Sub-Section--%%%%%%%%%%%%%%%%%%%%%%%%%%%%%%%%%%%%%%%%%%
\subsubsection{FS criterion}\label{subsubsec:evaluation_metrics_entropy}
Evaluating the quality of a multi-labeled clustering system is challenging. One of the challenges in evaluating multi-labeled clustering systems is that traditional measures, such as class and cluster Entropy\footnote{code can found at: \url{https://github.com/mkhadivi/ClusterEvaluation}}, do not consider multi-labeling. To overcome this, we used a novel measure called the FS criterion \cite{FS}. This criterion can be used to evaluate not only the performance of the whole clustering system but also the performance of each cluster or class.

In FS criterion\footnote{code can be found at: \url{https://github.com/cominsys/FS_criterion}}, results should be evaluated using a metric that reflects the "goodness" of the result. This metric is cluster FS. In addition, a metric is needed to evaluate the "compactness" of the results, called class FS. Suppose $S=\{s_1, s_2,..., s_{N_s}\}$ a set of samples, $C=\{c_1, c_2,..., c_{N_c}\}$ a set of true classes(given by an expert) and $\Omega=\{{\omega}_1, {\omega}_2,..., {\omega}_{N_{\Omega}}\}$ a set of given clusters(usually determined by a system), each sample $k$ in $S$ can be assigned to some of the classes in $C$ and the same sample $k$ can be assigned to some of the clusters in $\Omega$. The following metrics are defined.

\begin{itemize}
    \item \textbf{FS Cluster}\\
FS Cluster measures the degree of diversity or mixed membership within individual clusters. A lower cluster criterion value suggests that the cluster has a more homogeneous distribution of class labels or attributes, indicating well-defined and coherent clusters. Conversely, a higher cluster criterion value implies a more mixed or heterogeneous distribution of class labels or attributes within the cluster, indicating less well-defined clusters. So, the lower score is the best. Score of cluster ${\omega}_j$ is defined as:
\begin{equation}
    \label{eq: ClusterEntropyj}
        FS({\omega}_j) = \log{(\sum_{i=1}^{N_c}{{Sr}_{ij}})} - {\left(\sum_{i=1}^{N_c}{{Sr}_{ij}}\right)}^{-1} \times {\sum_{i=1}^{N_c}{({Sr}_{ij} \times \log{({Sr}_{ij})}})}
\end{equation}

where $N_C = \|C\|$ denotes number of classes and ${Sr}_{ij}$ is the Sum of the score of samples in cluster $j$ which are assigned to class $i$. Total cluster score for all clusters is defined as:

\begin{equation}
        \label{eq: ClusterEntropy}
        FS(\Omega) = 
        \sum_{j=1}^{N_{\Omega}}{\left(FS({\omega}_j) \times 
        \sum_{i=1}^{N_c}{{Sr}_{ij}} \times 
        {\left( \sum_{v=1}^{N_{\Omega}}{\sum_{u=1}^{N_c}{{Sr}_{uv}}}\right)}^{-1} \right)}
    \end{equation}

where $N_C = \|C\|$ and $N_{\Omega} = \|{\Omega}\|$ denote number of classes and clusters respectively.

\item \textbf{FS Class criterion}\\
Class criterion evaluates how well the created clusters represent data points of the same class. The optimal FS criterion value for a class is zero. If samples of the same class exist in several clusters, the class criterion score increases. Score of class $c_i$ is defined as:

\begin{equation}
        \label{eq: ClassEntropyi}
        FS(c_i) = 
        {\log{\sum_{j=1}^{N_{\Omega}}{{Ss}_{ij}}}} - 
        {\sum_{j=1}^{N_{\Omega}}{{Ss}_{ij}}}^{-1} \times 
        {\sum_{j=1}^{N_{\Omega}}{{{Ss}_{ij}} \times {\log{{Ss}_{ij}}}}}
\end{equation}

where $N_{\Omega} = \|{\Omega}\|$ denotes number of clusters and ${Ss}_{ij}$ is the Sum of the score of samples in class $i$ which are assigned to cluster $j$. Total class score for all classes is defined as:

\begin{equation}
    \label{eq: ClassEntropy}
    FS(C) = {\sum_{i=1}^{N_C}{\left( FS(c_i) \times {\sum_{j=1}^{N_{\Omega}}{{Ss}_{ij}}} \times {\left( \sum_{u=1}^{N_C}{\sum_{v=1}^{N_{\Omega}}{{Ss}_{uv}}} \right)}^{-1} \right)}}
\end{equation}
where $N_C = \|C\|$ and $N_{\Omega} = \|{\Omega}\|$ denote number of classes and clusters respectively.

It is important to note that these two metrics work in the opposite direction: reducing cluster criterion increases class criterion and vice versa. Therefore, a measurement criterion based on these two criteria will be useful. For this purpose, the arithmetic weighted average can be used. The total criterion is defined based on the following: 

    \begin{equation}
        \label{eq: Entropy}
       FS=\frac{w_{\Omega} \times FS(\Omega) + w_C  \times FS(C)}{w_{\Omega} + w_C}
    \end{equation}

where $w_{\Omega}$ is the corresponding cluster criterion weight, and $w_C$ is the standard class criterion weight. This proposed method uses an equal weight for both of them. Table \ref{tbl:results_entropy} shows this metric's results.
\end{itemize}

%%%%%%%%%%%%%%%%%%%%%%%%%%%%--Sub-Sub-Section--%%%%%%%%%%%%%%%%%%%%%%%%%%%%%%%%%%%%%%%%%%
\subsubsection{Topic Evaluation}\label{subsubsec:topic_evaluation}
To evaluate the correctness of the extracted threads, a metric is needed to compare the topics in the GT and the system. This paper uses Topic Precision and Topic Recall as performance evaluation scales. F-measure is a combination of Topic Precision and Topic Recall. All three of these criteria have been used to evaluate all windows.

\begin{itemize}
    \item \textbf{Topic Precision}\\
    The number of extracted Topics that match the Topics in GT. 
    \begin{equation}
        \label{eq: TopicPrecision}
       Topic\ Precision=\frac{\mid topics\ matches\ to\ GT\mid}{\mid extracted\ topics\mid}
    \end{equation}
    
    \item \textbf{Topic Recall}\\
    The number of correctly extracted Topics from GT Topics.
    \begin{equation}
        \label{eq: TopicRecall}
       Topic\ Recall=\frac{\mid successfully\ detected\ GT\ topics\mid}{\mid GT\ topics\mid}
    \end{equation}
    
    \item \textbf{Topic F1-Measure}\\
    Using the above concepts, F-measure is defined as follows
     \begin{equation}
        \label{eq: TopicF1}
       F_1-measure=2\times\frac{Topic\ Precision\times T o p i c\ Recall}{Topic\ Precision+Topic\ Recall}
    \end{equation}
\end{itemize}
These metrics results are presented in Table \ref{tbl:results_topic_evaluation}.

%%%%%%%%%%%%%%%%%%%%%%%%%%%%%%--Sub-Section--%%%%%%%%%%%%%%%%%%%%%%%%%%%%%%%%%%%%%%%%%%
\subsection{Results}\label{subsec:results}
\begin{figure}
    \centering
    \includegraphics[width=1.0\textwidth]{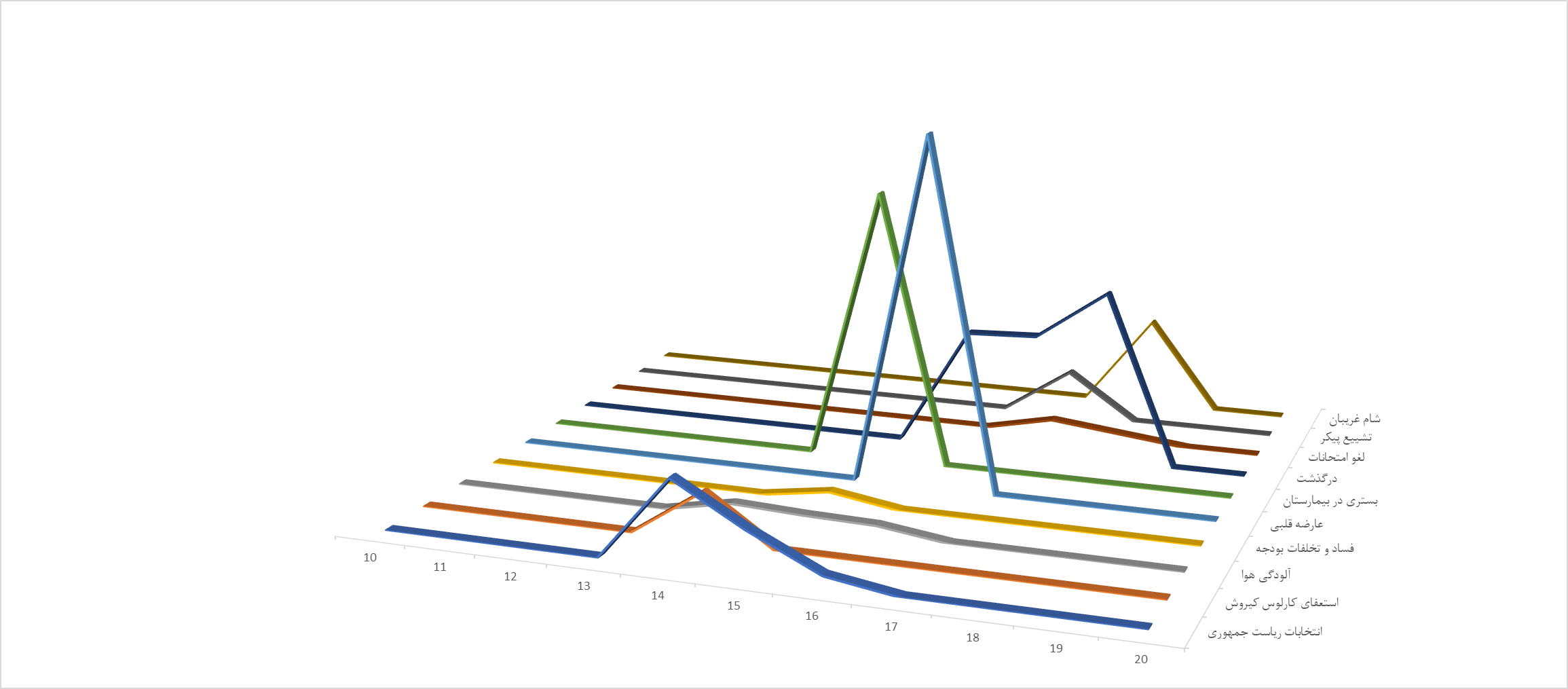}
    \caption{The trends of top topics generated by the proposed method in each batch.}
    \label{fig:Trends}
\end{figure}

For an overall analysis of the algorithm performance, the proposed framework demonstrated the best values in both FS criterion and Topic evaluation, as per Tables \ref{tbl:results_entropy} and \ref{tbl:results_topic_evaluation}. This work combines AGF with different graph embedding and clustering techniques. These tables show that combining AGF with the Node2Vec graph embedding method and the HDbscan clustering algorithm exhibits the best performance. Figure  \ref{fig:Trends} displays the progression of the top three topics in each window over the 'Hashemi death' super topic to visualize the trends. The topics this framework identifies appear on the left of the charts and align with the right-coming topics, which constitute Ground Truth. The temporal trend of these topics and how they change over time can be observed. As the plot indicates, window 16 is the change window wherein the event's first sub-topic, 'Heart attack', emerged.

%%%%%%%%%%%%%%%%%%%%%%%%%%%%--Sub-Sub-Section--%%%%%%%%%%%%%%%%%%%%%%%%%%%%%%%%%%%%%%%%%%
%\subsubsection{Entropy}\label{subsubsec:results_entropy}

\begin{table}[]
    \centering
    \caption{The proposed method's FS criterion results compared to other methods.}
    \label{tbl:results_entropy}
    \begin{tabular}{cccc}
        \hline
        \multirow{2}{*}{\textbf{Method}}  &\multicolumn{3}{c}{\textbf{FS criterion}} \\
            &Cluster    &Class   &Total  \\
        %\cline{2-11}
        \hline
        
        AGF + Node2Vec+ HDbscan (proposed)   & 1.328505 & 0.690471 & \textbf{1.009488202}  \\
        AGF + deep walk + HDbscan (proposed) & 1.594577 & 0.645277 & 1.119927096 \\
        AGF + Node2Vec+ kmeans (proposed)    & 1.402442 & 0.750816 & 1.076629029  \\
        AGF + deep walk + kmeans (proposed)  & 1.470772 & 1.273735 & 1.372253661  \\
        AGF(base) \cite{AGF_base}            & 1.530920 & 0.738440 & 1.134680404  \\
        Twevent \cite{twevent}               & 1.721218 & 0.388564 & 1.05489122   \\
        HUPM \cite{HUPM}                     & 1.729917 & 0.362058 & 1.045987555  \\
        HUPC \cite{HUPC}                     & 1.520165 & 0.825380 & 1.172772534  \\
       % Word2Vec + OPTICS                    & 1.352988759 & 1.219940783 & 1.286464771 \\
        %FastText + OPTICS                    & 1.46328151  & 1.141669102 & 1.302475306 \\
        %GloVe + Kmeans                    & 1.805302425 & 1.028225611 & 1.416764018 \\
        \hline
    \end{tabular}
\end{table}

%%%%%%%%%%%%%%%%%%%%%%%%%%%%--Sub-Sub-Section--%%%%%%%%%%%%%%%%%%%%%%%%%%%%%%%%%%%%%%%%%%
%\subsubsection{Topic Evaluation}\label{subsubsec:results_topic_evaluation}

\begin{table}[]
    \centering
    \caption{The Topic Evaluation results of the proposed method compared to other methods.}
    \label{tbl:results_topic_evaluation}
    \begin{tabular}{cccc}
        \hline
        \multirow{2}{*}{\textbf{Method}}  &\multicolumn{3}{c}{\textbf{Topic Evaluation}} \\
            &Precision    &Recall   &F-Measure  \\
        %\cline{2-11}
        \hline
        AGF + Node2Vec+ HDbscan (proposed)   & 0.825926 & 0.510293 & \textbf{0.630831}  \\
        AGF + deep walk + HDbscan (proposed) & 0.774018 & 0.507079 & 0.612738 \\
        AGF + Node2Vec+ kmeans (proposed)    & 0.825555 & 0.480134 & 0.607154 \\
        AGF + deep walk + kmeans (proposed)  & 0.818182 & 0.465517 & 0.593407 \\
        AGF (base) \cite{AGF_base}           & 0.497221 & 0.819564 & 0.618938 \\
        TweEvent \cite{twevent}              & 0.620762 & 0.482383 & 0.542894 \\
        HUPM \cite{HUPM}                     & 0.895330 & 0.299086 & 0.448387 \\
        HUPC \cite{HUPC}                     & 0.941176 & 0.172414 & 0.291439  \\
        %Word2Vec + OPTICS                &   &   &   \\
        %FastText + OPTICS                &  &  &  \\
        %GloVe + Kmeans                   &   &   &   \\

        \hline
    \end{tabular}
\end{table}

%%%%%%%%%%%%%%%%%%%%%%%%%%%%%%%%--Section--%%%%%%%%%%%%%%%%%%%%%%%%%%%%%%%%%%%%%%%%%%%%
\section{Conclusion}\label{sec:conclusion}
This paper proposes a trending topic detection framework utilizing HWA combined with graph embedding techniques and clustering methods. This framework is applied to social media posts collected from Telegram. Initially, the co-occurrence of words in posts is extracted. Then, the calculations are performed using HWA, leading to generating the AGF graph. This graph is then input into a graph embedding algorithm. Finally, after dimension reduction, the vectors obtained are clustered together to group similar topics. This approach is applied to a Persian language dataset collected from Telegram, and several experimental studies are conducted to assess its performance. The experimental results indicate that this framework outperforms other prevalent topic detection approaches.
%%%%%%%%%%%%%%%%%%%%%%%%%%%%%%%%--Section--%%%%%%%%%%%%%%%%%%%%%%%%%%%%%%%%%%%%%%%%%%%%
\section*{Declarations}
\subsection*{Funding}
No funds, grants, or other support were received.
\subsection*{Conflict of interest}
The authors have no conflicts of interest to declare that are relevant to the content of this article.
\subsection*{Availability of data and materials}
The dataset generated during the current study, Sep\_TD\_Tel01, is available in the Mendeley repository, \url{https://doi.org/10.17632/372rnwf9pc}.
\subsection*{Authors' contributions}
\textbf{Mehrdad Ranjbar-Khadivi:} Conceptualization, Methodology, Investigation, Data curation, Validation, Writing - original draft. \textbf{Mohammad-Reza Feizi-Derakhshi:} Project administration, Formal analysis, Supervision. \textbf{Shahin Akbarpour:} Supervision. \textbf{Babak Anari:} Supervision.
\subsection*{Ethics approval}
Not applicable.
%%%%%%%%%%%%%%%%%%%%%%%%%%%%%%%%--Section--%%%%%%%%%%%%%%%%%%%%%%%%%%%%%%%%%%%%%%%%%%%%
% \begin{appendices}
% \end{appendices}


\bibliography{sn-bibliography}% common bib file


\end{document}